\documentclass{article}

\usepackage{PRIMEarxiv}

\usepackage[utf8]{inputenc} % allow utf-8 input
\usepackage[T1]{fontenc}    % use 8-bit T1 fonts
\usepackage{hyperref}       % hyperlinks
\usepackage{url}            % simple URL typesetting
\usepackage{booktabs}       % professional-quality tables
\usepackage{amsfonts}       % blackboard math symbols
\usepackage{nicefrac}       % compact symbols for 1/2, etc.
\usepackage{microtype}      % microtypography
\usepackage{lipsum}
\usepackage{fancyhdr}       % header
\usepackage{graphicx}       % graphics
\graphicspath{{media/}}     % organize your images and other figures under media/ folder

\RequirePackage{algorithm}
\RequirePackage{algorithmic}
\usepackage{mathtools}
\usepackage{amsthm}
\usepackage{xcolor}
\newcommand{\mycolor}{black}
\usepackage{amsmath, amssymb,bm}
\usepackage{amsfonts}
\usepackage{bbm}

\usepackage{subfigure}
\usepackage{lscape}

\theoremstyle{plain}
\newtheorem{theorem}{Theorem}[section]

\theoremstyle{definition}

\theoremstyle{remark}

\title{\scalebox{0.9}{Neural Operators Meet Energy-based Theory:}\\
\scalebox{0.9}{Operator Learning for Hamiltonian and Dissipative PDEs}
\thanks{\textit{\underline{Citation}}: 
\textbf{Authors. Title. Pages.... DOI:000000/11111.}} 
}

\author{%
 Yusuke Tanaka\textsuperscript{$\ast$},\,
 Takaharu Yaguchi\textsuperscript{$\dagger$},\,
 Tomoharu Iwata\textsuperscript{$\ast$},\,
 Naonori Ueda\textsuperscript{$\ast,\ddagger$}
 \and
 \textsuperscript{$\ast$}{NTT Communication Science Laboratories},\,
 \textsuperscript{$\dagger$}{Kobe University},\,
 \textsuperscript{$\ddagger$}{RIKEN Center for AIP}
 \and
 \texttt{\{ysk.tanaka,tomoharu.iwata\}@ntt.com},\,
 \texttt{yaguchi@pearl.kobe-u.ac.jp},\,
 \texttt{naonori.ueda@riken.jp}
}

\begin{document}
\maketitle

\begin{abstract}
The operator learning has received significant attention in recent years, with the aim of learning a mapping between function spaces. Prior works have proposed deep neural networks (DNNs) for learning such a mapping, enabling the learning of solution operators of partial differential equations (PDEs). However, these works still struggle to learn dynamics that obeys the laws of physics. This paper proposes {\em Energy-consistent Neural Operators (ENOs)}, a general framework for learning solution operators of PDEs that follows the energy conservation or dissipation law from observed solution trajectories. We introduce a novel penalty function inspired by {\em the energy-based theory of physics} for training, in which the energy functional is modeled by another DNN, allowing one to bias the outputs of the DNN-based solution operators to ensure energetic consistency without explicit PDEs. Experiments on multiple physical systems show that ENO outperforms existing DNN models in predicting solutions from data, especially in super-resolution settings.
\end{abstract}

% keywords can be removed

\section{Introduction}
\label{sec:introduction}
Many physical systems are described by partial differential equations (PDEs)~\cite{evans:partial}.
\color{\mycolor}
Obtaining their solutions
\color{black}
is fundamental in such disciplines as weather forecasting~\cite{peter:origins}, molecular modeling~\cite{stoltz:partial}, astronomical simulations~\cite{richard:partial}, and jet engine design~\cite{michael:parametric}.
Traditionally, one crafts a PDE by hand and 
\color{\mycolor}
obtains its solution via a numerical simulation.
\color{black}
However, 
designing the PDEs requires domain knowledge and great effort; 
\color{\mycolor}
the numerical simulation depends on a spatio-temporal mesh; moreover, it is often high-resolution, yielding high computational costs.
\color{black}

In the machine learning community, interest in a novel approach called {\em operator learning} is growing,
which
\color{\mycolor}
predicts mesh-free solutions from data; PDE designs and numerical simulations are not required.
\color{black}
Generally speaking, the goal of operator learning is to obtain a mapping (i.e., {\em an operator}) between function spaces.
In a setting of differential equations, the solution operator is the mapping from the input function (e.g., initial and boundary conditions) to the output solution function.
There have been many deep neural networks (DNNs) for operator learning, such as deep operator networks (DeepONets)~\cite{lu2021learning} and Fourier neural operators (FNOs)~\cite{li2021fourier}, which can be used for approximating a solution operator from many pairs of input and output functions.
However, these works assume the availability of a large amount of training data with high spatial and temporal resolutions.
In practice, obtaining such data is quite costly; these existing works struggle to accurately predict solutions from low-resolution data.

One promising approach is using the prior knowledge of physics as an inductive bias for training DNNs, which has recently been addressed under the name of {\em physics-informed machine learning (PIML)}~\footnote{%
\color{\mycolor}
Note that {\em PIML} refers to the field of machine learning for exploiting various forms of physics knowledge and is distinct from {\em PINNs} that directly use PDEs, which is one approach in PIML.
\color{black}
}~\cite{karniadakis:physics,Chuizheng:when}.
A few works have introduced physics-informed inductive bias into the operator learning framework, called physics-informed DeepONets (PI-DeepONets)~\cite{sifan:learning} and physics-informed neural operators (PINOs)~\cite{li2023:physics,sifan:learning}, by adding PDE constraints to the loss function, as in physics-informed neural networks (PINNs)~\cite{raissi:physics}.
However, these works require explicit PDEs for training DNNs and cannot be applied to physical systems for which 
\color{\mycolor}
their PDEs
\color{black}
are unknown.

The Hamiltonian neural network (HNN)~\cite{greydanus:hamiltonian} and its variants (e.g.,~\cite{chen:symp,zhong:symplectic,matsubara:deep}) have introduced an inductive bias based on the Hamiltonian mechanics, 
which can be used for inferring dynamics that follows the basic laws of physics, such as energy conservation or dissipation, without explicit differential equations.
However, these works aim to obtain the time-evolution at a point of time, not solution operators, from data; hence they require computationally expensive numerical simulations to obtain solutions.

This paper proposes {\em Energy-consistent Neural Operators (ENOs)}, a general 
\color{\mycolor}
data-driven
\color{black}
framework for learning solution operators of 
\color{\mycolor}
{\em hidden}
\color{black}
PDEs that adhere to the laws of physics, such as energy conservation or dissipation,
without explicit PDEs.
Our proposed framework assumes that the solution operators are parameterized by a DNN
\color{\mycolor}
(called {\em operator net})
\color{black}
, where it does not depend on a particular choice of the DNN architecture of solution operators as long as it is differentiable.
The most significant contribution of ENO is a novel penalty function inspired by the energy-based theory~\cite{furihata:finite,quispel:solving}, which allows one to bias the time-derivative of solution functions to ensure the energy conservation or dissipation law.

To obtain the penalty function, we model an {\em unknown} system's total energy that is defined by a {\em functional} using another DNN 
\color{\mycolor}
(called {\em energy net})
\color{black}
; we then derive a gradient flow of the energy functional by calculating the functional derivatives (also called variational derivatives) via automatic differentiation.
%Here, this gradient flow is the natural gradient on symplectic or Riemannian manifolds; 
\color{\mycolor}
According to the energy-based theory,
\color{black}
the energetic consistency is guaranteed by employing this gradient flow as the time-derivative of the solution.
Our penalty encourages the time-derivative of the DNN-based operator to be equal to the gradient flow of the energy functional.
\color{\mycolor}
In training, the operator net and energy net are simultaneously optimized by minimizing the data loss to predict solutions and the penalty.
The important thing here is that ENO estimates not only the solution operator (i.e., operator net) but also the energy functional (i.e., energy net); hence, it does not need the explicit PDEs and can obtain solution operators of hidden PDEs, unlike PI-DeepONets and PINOs.
\color{black}
In testing, efficient and mesh-free simulation of physical systems is possible using the learned solution operator.

One advantage of ENO is that it can consider the penalty term at arbitrary points not included in the training data, 
\color{\mycolor}
yielding a smoothing effect based on the laws of physics over the entire spatio-temporal domain.
This property is especially helpful in super-resolution settings: we predict higher-resolution data from only lower-resolution data.
\color{black}

%\vspace{-0.5mm}
The following are the main contributions of our work:
\begin{itemize}
 %\vspace{-2.5mm}
 %\setlength{\leftskip}{-1mm}
%\setlength{\parskip}{-1.3mm}
 %\setlength{\itemsep}{-0.6mm}
 \color{\mycolor}
 \item Energy-consistent Neural Operator (ENO) is the first framework based on the energy-based theory for learning solution operators of Hamiltonian and dissipative PDEs from data without explicit PDEs.
 \item We propose a penalty function designed to encourage the time-derivative of DNN-based operators to align with the gradient flow of the energy functional.
 \color{black}
 \item Experiments on multiple systems show that ENO more accurately predicts higher-resolution solutions while ensuring energetic consistency than baselines, especially when the training data is lower-resolution.
\end{itemize}

\section{Related Work}
\label{sec:related}
Roughly speaking, this work bridges two research topics in the machine learning community, that is, neural networks for operator learning and 
\color{\mycolor}
the energy-based theory of physics 
\color{black}
as an inductive bias.
Below we describe prior works that represent two research lines.
Table~\ref{tb:comparison} compares the proposed method with the existing representative methods.
\begin{table*}[t]
 %\vspace{-2mm}
 \centering
 \caption{Comparison of proposed method (ENO) with existing methods: we compared the methods with six items.
 First and second items represent that they can handle physical systems represented by (A) ODEs and (B) PDEs.
 Third item (C) states that they can estimate solution operators; methods without a check mark require numerical simulations to obtain solutions.
 The fourth and fifth items represent that the methods can use the physics priors, i.e., (D) energy conservation law and (E) energy dissipation law, respectively, as an inductive bias for training.
 Last item (F) shows that training can be done from only observed data; method (i.e., PI-DeepONet and PINO) without a check mark assumes that explicit PDEs are known.
 }
 %\vskip 0.01in
 \small
 \begin{sc}
 \begin{tabular}{ l c c c c c} \toprule
  &DeepONet / FNO &PI-DeepONet / PINO &HNN &DGNet &ENO \\
  %&\cite{lu2021learning,li2021fourier} &\cite{li2023:physics,sifan:learning}
  %&\cite{greydanus:hamiltonian} &\cite{matsubara:deep} & \\
  \midrule
  (a) ODE systems &$\surd$ &$\surd$ &$\surd$ &$\surd$ &$\surd$ \\[2pt]
  (b) PDE systems &$\surd$ &$\surd$ & &$\surd$ &$\surd$  \\[2pt]
  (c) Solution operator &$\surd$ &$\surd$ & & &$\surd$ \\[2pt]
  (d) Energy conservation law & &$\surd$ &$\surd$ &$\surd$ &$\surd$ \\[2pt]
  (e) Energy dissipation law & &$\surd$ & &$\surd$ &$\surd$ \\[2pt]
  (f) Without explicit equations &$\surd$ & &$\surd$ &$\surd$ &$\surd$ 
  \\
  \bottomrule
 \end{tabular}
 \end{sc}
 %}
 \label{tb:comparison}
 %\vskip -0.15in
\end{table*}

\paragraph{Neural Network Models for Operator Learning}
Many deep neural network (DNN) models have been proposed
for learning solution operators of PDEs~\cite{bhattacharya:model,ravi:physics,li2020:multipole,rahman2022:uno,cao:choose,de:generic,poli:transform,gupta:multi}.
In these studies, a mapping between input functions (e.g., initial conditions) and output functions (i.e., solutions) is parameterized by DNNs.
Graph neural operators construct solution functions as a convolution of input function values on a graph-based discretization using DNN-based kernels~\cite{li2020:neural}.
Fourier neural operators (FNOs), which are their extensions, allow for efficiently computing the kernels by introducing Fourier approximations in the frequency domain~\cite{li2021fourier}.
\color{\mycolor}
%The FNO is often used to predict solution functions at each discrete time step.
Deep operator networks (DeepONets) are one of the most widely-used architectures for operator learning;
%Meanwhile, deep operator networks (DeepONets) are more general architectures for predicting solution functions defined on a spatio-temporal domain; 
\color{black}
the architecture design is motivated by the universal approximation theorem for operators~\cite{lu2021learning}.

These models are computationally efficient for testing.
For instance, in the case of an initial value problem, obtaining a 
\color{\mycolor}
mesh-free solution 
\color{black}
for an unseen initial condition requires only a forward pass of the learned DNN. 
\color{\mycolor}
This drastically reduces the computational cost associated with numerical simulations. 
Nevertheless, their performance is limited in situations where there is an insufficient amount of training data with high spatial and temporal resolutions. 
\color{black}

Physics-informed DeepONets (PI-DeepONets)~\cite{sifan:learning} and physics-informed neural operators (PINOs)~\cite{li2023:physics} have tackled the data scarcity problem by introducing PDE constraints, in a manner similar to physics-informed neural networks~\cite{raissi:physics}.
PDE constraints are expected to infer the dynamics that follows the laws of physics even with limited data.
These works, however, need explicit PDEs for training: PI-DeepONets and PINOs are inapplicable to our problem setting, where we assume that explicit PDEs are unknown.

\paragraph{
\color{\mycolor}
Energy-based Theory of Physics as an Inductive Bias}
An inductive bias is helpful for training models from a limited amount of data.
Recently, many machine learning models utilize the prior knowledge of Hamiltonian mechanics~\cite{goldstein:mechanics} as an inductive bias for inferring the dynamics that ensures the energetic consistency of physics~\cite{chen:symplectic,chen:symp,marc:simplifying,zhong:symplectic,rath:symplectic,tanaka:symplectic}.
\color{\mycolor}
Here, Hamiltonian mechanics can be regarded as a particular case of the energy-based theory~\cite{celledoni:preserving}.
\color{black}

The pioneering work is the Hamiltonian neural network (HNN)~\cite{greydanus:hamiltonian}, of which the key idea is parameterizing the Hamiltonian (i.e., energy) using DNNs; then, the time evolution of the ordinary differential equation (ODE) systems is given by 
\color{\mycolor}
the energy gradient (called symplectic gradient).
\color{black}
The energy conservation law is guaranteed by employing the symplectic gradient as the time-derivative.
By estimating the Hamiltonian from data, the HNN makes it possible to embed energy conservation laws into the network architectures without explicit differential equations.

Most recent works have expanded the scope of application, such as Hamiltonian systems with energy dissipation~\cite{zhong:dissipative,sosanya:dissipative}, Hamiltonian systems with controllable inputs~\cite{desai:port}, stiff Hamiltonian systems~\cite{liang:stiffness}, odd-dimensional chaotic systems~\cite{kevin:weak}, and Poisson systems~\cite{jin:learning}.
\color{\mycolor}
The DGNet~\cite{matsubara:deep} has been extended to handle Hamiltonian and dissipative PDEs by employing the energy-based theory~\cite{furihata:finite,quispel:solving}.
\color{black}
However, since all of the existing models infer the time-evolution at a point of time, not solution operators, from data, they require numerical simulations to obtain solutions. 
This approach has a significant disadvantage in terms of computational costs, especially in the PDE setting.
\color{\mycolor}
In addition, predictions can only be made for a predefined spatio-temporal discretization when applied to PDEs~\cite{matsubara:deep,eidnes:pseudo}.
\color{black}

\paragraph{Our Work}
We integrate the energy-based theory into the operator learning framework, which allows one to utilize the physics prior (i.e., energy conservation and dissipation laws) to obtain solution operators without explicit PDEs.
It should be noted that our problem setting is different from that of PI-DeepONets and PINOs, where explicit PDEs are assumed to be known.
\color{\mycolor}
In addition, unlike the HNN and its extensions that require computationally expensive numerical simulations, the learned operators can be used for efficient simulations without discretization.
\color{black}

\section{Preliminary}
\subsection{Energy-based Theory of physics}
\label{sec:energy}
This section presents an overview of the energy-based theory~\cite{furihata:finite,quispel:solving}, a mathematical framework that generalizes
\color{\mycolor}
Hamiltonian mechanics.
The energy-based theory enables one to handle Hamiltonian and dissipative partial differential equations (PDEs) as well as Hamiltonian systems (defined by ordinary differential equations)~\cite{celledoni:preserving}.
\color{black}

\begin{figure}[t]
    \centering
    %\vskip -0.02in
    \includegraphics[width=60mm]{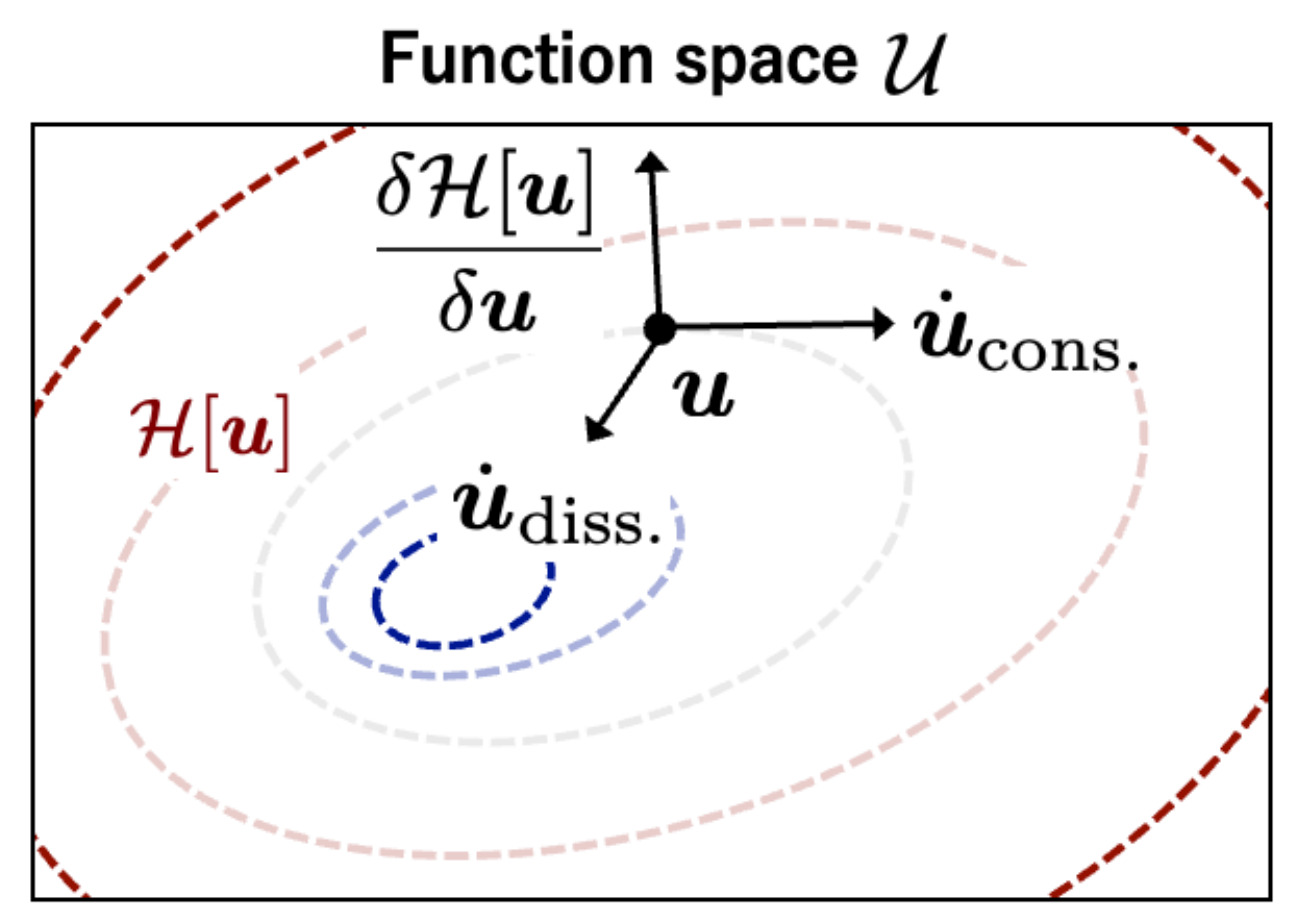}
    %\vskip -0.15in
    \caption{
    \color{\mycolor}
    Intuitive view of gradient flows~\eqref{eq:H_pde}.
    Dashed lines represent contours of energy functional $\mathcal{H}$ on function space $\mathcal{U}$; blue dashed line represents low energy.
    Functional derivative $\delta\mathcal{H}/\delta\bm{u}$ is orthogonal to the contour at $\bm{u}$.
    Systems follow a flow $\dot{\bm{u}}_{\rm{cons.}}$ conserving $\mathcal{H}$ if $\mathcal{G}$ is skew-symmetric and a flow $\dot{\bm{u}}_{\rm{diss.}}$ dissipating $\mathcal{H}$ if $\mathcal{G}$ is negative (semi) definite.
    \color{black}
    }
    %\vskip -0.15in
    \label{fig:gradient}
\end{figure}

Let $\mathcal{T}=\mathbb{R}_{\geq 0}$ be the time domain, and let $\mathcal{X}\subset\mathbb{R}^D$ be the $D$-dimensional bounded spatial domain.
We consider physical systems defined on the time-space $(t,\bm{x})\in\mathcal{T} \times \mathcal{X}$. 
Let $\bm{u}:\mathcal{T}\times\mathcal{X}\to\mathbb{R}^M$ be a solution function that represents the system's state, 
\color{\mycolor}
where $\bm{u}$ belongs to a function space $\mathcal{U}$.
\color{black}
In the energy-based formulation, the starting point is to define {\em energy functional} 
\color{\mycolor}
$\mathcal{H}:\mathcal{U}\to\mathbb{R}$, 
\color{black}
which denotes the system's total energy.
The energy functional is given by
\begin{align}
    \mathcal{H}[\bm{u}]
        &= \int_{\mathcal{X}} F(\bm{u},\partial_{\bm{x}}\bm{u},\partial_{\bm{xx}}\bm{u},\ldots) \,d\bm{x},
        \label{eq:energy}
\end{align}
where 
\color{\mycolor}
$F:\mathcal{U}\to\mathbb{R}$
\color{black}
is called an energy density.
One can observe that energy functional $\mathcal{H}$ is obtained by integrating density $F$ over spatial domain $\mathcal{X}$.
Here, we adopt shorthand notation $\partial_{\bm{x}}\bm{u}, \partial_{\bm{xx}}\bm{u},\ldots$ for partial derivatives $\partial \bm{u}/\partial\bm{x}, \partial^2 \bm{u}/\partial\bm{x}^2$, and so on.
Traditionally, the energy density $F$ is manually designed to suit the system.
Given the energy functional $\mathcal{H}$, PDE dynamics is given by
\begin{align}
    \dot{\bm{u}}
        &= \mathcal{G} \frac{\delta \mathcal{H}[\bm{u}]} {\delta \bm{u}},
    \label{eq:H_pde}
\end{align}
where $\dot{\bm{u}}$ denotes $\partial \bm{u}/\partial t$,
\color{\mycolor}
and the right-hand side of~\eqref{eq:H_pde} defines {\em the gradient flow of the energy functional}.
Here, $\mathcal{G}$ is typically a constant linear differential operator with respect to $\bm{x}$; the specific form of $\mathcal{G}$ depends on the class of systems, which we detailed in the following paragraph.
$\delta \mathcal{H}/\delta \bm{u}$ is a functional derivative (also called a variational derivative) of $\mathcal{H}$, which denotes a change in functional $\mathcal{H}$ to a change in function $\bm{u}$.
Fig.~\ref{fig:gradient} shows an intuitive view of the gradient flow~\eqref{eq:H_pde}.
\color{black}
\color{\mycolor}
Importantly, the energetic behavior of~\eqref{eq:H_pde} depends on the choice of $\mathcal{G}$, and the following theorem is known to hold~\cite{celledoni:preserving}.
\begin{theorem}{\bf (Energy conservation and dissipation)}
\label{thm:theorem}
The system~\eqref{eq:H_pde} follows the energy conservation law if $\mathcal{G}$ is a skew-symmetric operator and the energy dissipation law if $\mathcal{G}$ is a negative (semi) definite operator.
%and hence $\int_{\mathcal{X}} \bm{u}\mathcal{G}\bm{u}\,d\bm{x}=0$, for any $\bm{u}$, then the system~\eqref{eq:H_pde} follows an energy conservation law.
\end{theorem}
%\begin{theorem}{\bf (Energy dissipation)}
%\label{thm:theorem2}
%If $\mathcal{G}$ is a constant negative (semi) definite operator, and hence $\int_{\mathcal{X}} \bm{u}\mathcal{G}\bm{u}\,d\bm{x}\leq0$, for any $\bm{u}$, then the system~\eqref{eq:H_pde} follows an energy dissipation law.
%\end{theorem}
According to~\cite{olver:applications}, the functional derivative in~\eqref{eq:H_pde} can be calculated as follows:
\begin{align}
    \frac{\delta\mathcal{H}[\bm{u}]}{\delta u_m}
        &= \frac{\partial F}{\partial u_m}
        -\sum_{d=1}^D 
        \left[
            \frac{\partial}{\partial x_d}
            \left(
                \frac{\partial F}{\partial u_{m,d}}
            \right)
        \right]
        +\cdots,
    \label{eq:functional}
\end{align}
for $m=1,\ldots,M$.
Here, $u_m$ is the $m$th element in $\bm{u}$, $x_d$ is the $d$th element in $\bm{x}$, and $u_{m,d}$ denotes $\partial u_m/\partial x_d$.
\color{black}

\paragraph{\color{\mycolor}
Specific Form of $\mathcal{G}$
\color{black}
}
Differential operator $\mathcal{G}$ in~\eqref{eq:H_pde} depends on the class of systems; for example, the differential operator for energy-conserving systems (called {\em Hamiltonian PDEs}), such as the Korteweg--de Vries (KdV) equation, the advection equation, and the Burgers equation, is given by $\mathcal{G}=\partial/\partial \bm{x}$.
\color{\mycolor}
Energy-dissipating systems (called {\em dissipative PDEs}) include the Allen–Cahn equation, the Cahn-Hillard equation, and so on.
For example, the differential operator for the Cahn-Hillard equation is given by $\mathcal{G}=\partial^2/\partial \bm{x}^2$.

\paragraph{Mass Conservation}
In addition to the energy conservation or dissipation law, the system~\eqref{eq:H_pde} under the periodic boundary condition admits the mass conservation laws $\frac{\partial}{\partial t}\int \bm{u} dx = 0$ if $\mathcal{G}=\partial^p/\partial\bm{x}^p$, where $p\in\mathbb{N}$, which follows from the equation: $\frac{\partial}{\partial t}\int \bm{u} dx = \int \mathcal{G} \frac{\delta \mathcal{H}}{\delta \bm{u}}dx=0$.
\color{black}

The energy-based theory covers many other physical systems. 
For example, the Schr\"{o}dinger and the Ginzburg--Landau equations can be expressed by introducing complex state variables.
See~\cite{furihata:finite} for details.

\paragraph{Relation to Hamiltonian systems}
Hamiltonian systems are well known as ordinary differential equation (ODE) systems with energy conservation laws and can be regarded as a special case of~\eqref{eq:H_pde}.
In Hamiltonian mechanics~\cite{goldstein:mechanics}, the system's state $\bm{u}:\mathcal{T}\to\mathbb{R}^M$ is defined on the product space of generalized coordinates and generalized momenta.
The functional $\mathcal{H}$ and its functional derivative in~\eqref{eq:H_pde} are replaced with the energy {\em function} (called {\em Hamiltonian}) and its gradient, respectively.
The differential operator $\mathcal{G}$ is reduced to the skew-symmetric matrix.

\subsection{Operator Learning}
\label{sec:operator}
This section describes the operator learning framework.
Generally speaking, the aim of operator learning is to obtain a mapping between two infinite-dimensional function spaces from a finite set of observed input-output pairs. 
In the following, we elaborate on the problem of learning solution operators of PDE systems. 
Note that this problem setting includes the case of ODE systems as a special case.

\paragraph{Learning Solution Operators}
Let $\mathcal{A}$ and $\mathcal{U}$ be input and output function spaces.
Input function $\bm{a}\in\mathcal{A}$ corresponds to the initial or boundary conditions, constant or variable coefficients, forcing terms, and so on; the input function can be chosen freely, depending on what we want to generalize to.
Output function $\bm{u}\in\mathcal{U}$ corresponds to the solution, given input function $\bm{a}$.
The goal is to approximate a {\em solution operator} $\mathcal{S}:\mathcal{A}\to\mathcal{U}$, which is a non-linear map between the input and output function spaces.
Accordingly, the input and output functions satisfy the following relationship:
\begin{align}
    \bm{u} = \mathcal{S}[\bm{a}].
\end{align}
In the training phase, we assume that we have $I$ samples of input-output function pairs 
\color{\mycolor}
$\{(\bar{\bm{a}}_i,\bar{\bm{u}}_i)\mid i=1,\ldots,I\}$, where the superscript bar indicates point-wise observations for functions; namely, $\bar{\bm{a}}_i$ and $\bar{\bm{u}}_i$ are the finite sets of the input and output function values evaluated on the discretization points.
\color{black}
Let $\mathcal{Y}=\mathcal{T}\times\mathcal{X}$ denote the spatio-temporal domain, and let $\bm{y}\in\mathcal{Y}$ denote a {\em query point}.
Let $\bar{\mathcal{Y}}_i=\{\bm{y}_{i,j}\mid j=1,\ldots,J_i\}\subset\mathcal{Y}$ be a $J_i$-point discretization for the solution $\bm{u}_i$.
%Let $\bar{\mathcal{Y}}=\{\bm{y}_{j}\mid j=1,\ldots,J\}\subset\mathcal{Y}$ be a $J$-point discretization~\footnote{For simplicity, we assume identical discretization for all the observations. Our method can also apply to the case of multiple discretizations that differ for each index $i$ by adopting appropriate solution operators (e.g.,~\cite{li2021fourier}).} for solution $\bm{u}$.
Also, the input function $\bm{a}_i$ may have a different discretization for each index $i$.
Given the data, we wish to obtain solution operator $\bm{u}(\bm{y})=\mathcal{S}_{\bm{\theta}}[\bar{\bm{a}}](\bm{y})$ approximated by deep neural networks (DNNs), where $\bm{\theta}$ is their parameters.
The parameters can be estimated by minimizing the following mean squared error,
\begin{align}
    L(\bm{\theta})
    = \frac{1}{I} \sum_{i=1}^I \left(
    \frac{1}{J_i} \sum_{j=1}^{J_i}
    \Bigl\| \bar{\bm{u}}_i(\bm{y}_{i,j}) - \mathcal{S}_{\bm{\theta}}[\bar{\bm{a}}_i](\bm{y}_{i,j}) \Bigr\|^2
    \right),
    \label{eq:data_loss}
\end{align}
where $\bar{\bm{u}}_i(\bm{y}_{i,j})$ and $\mathcal{S}_{\bm{\theta}}[\bar{\bm{a}}_i](\bm{y}_{i,j})$ denote the observed and predicted solution values at query point $\bm{y}_{i,j}$ and $\|\cdot\|$ is the Euclidean norm.
In the test phase, given {\em unseen} input function $\bar{\bm{a}}^\ast$, the solution function $\bm{u}^\ast(\bm{y})$ can be predicted using the estimated operator, as $\bm{u}^\ast(\bm{y}) = \mathcal{S}_{\bm{\theta}}[\bar{\bm{a}}^\ast](\bm{y})$.
Note that $\mathcal{S}_{\bm{\theta}}[\bar{\bm{a}}^*](\bm{y})$ can predict a solution value for arbitrary query point $\bm{y}\in\mathcal{Y}$, potentially $\bm{y}\notin\bar{\mathcal{Y}}_1\cup\cdots\cup\bar{\mathcal{Y}}_I$.

\section{Energy-consistent Neural Operators}
\label{sec:eno}
\begin{figure}[t]
    \centering
    \includegraphics[width=65mm]{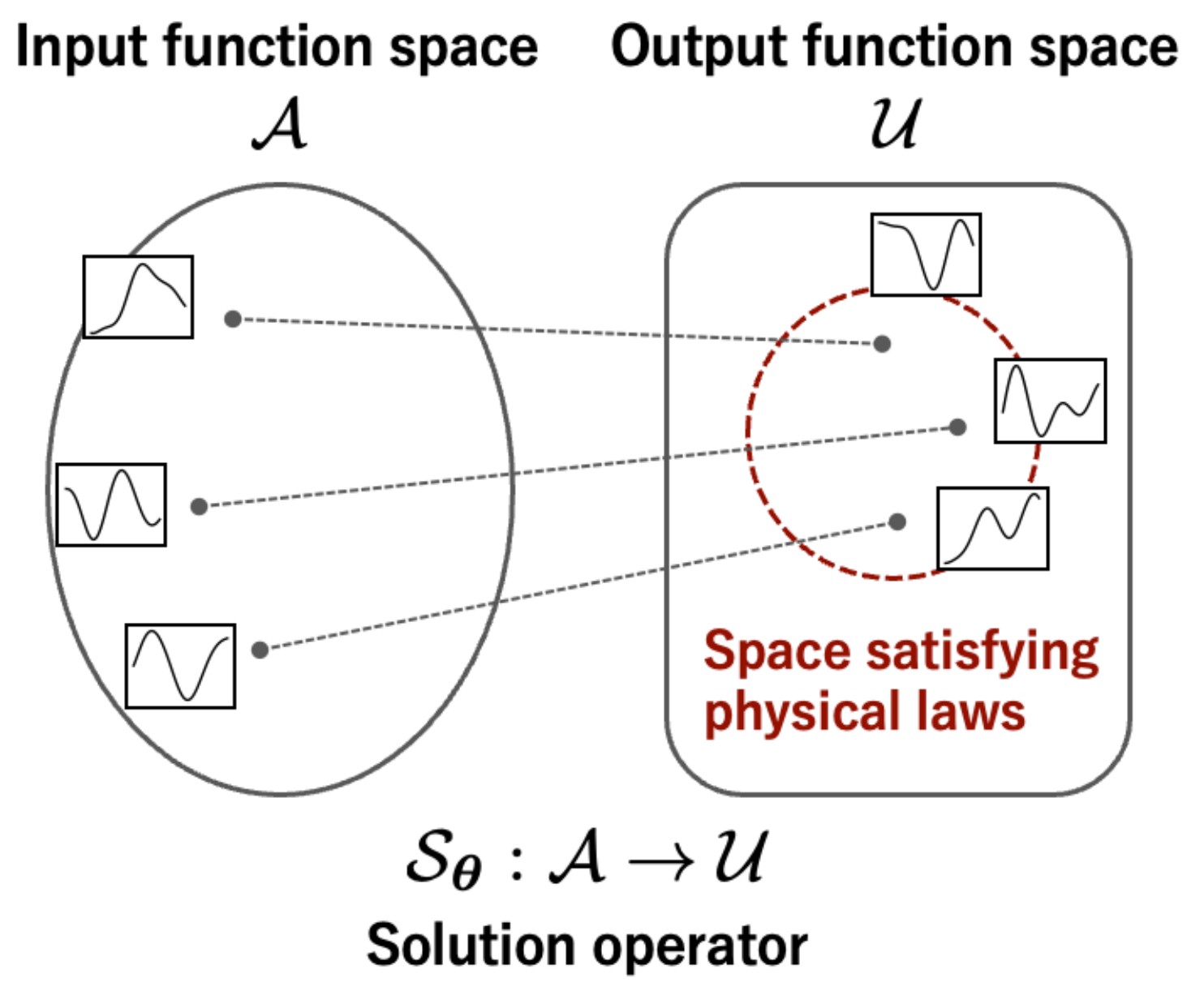}
    %\vskip -0.15in
    \caption{Basic idea of our proposed approach.
    Input functions are initial, boundary conditions, etc; output functions are the corresponding solutions.
    Goal is to obtain solution operator from input-output function pairs.
    Our aim is to introduce inductive biases such that the solution operator's output satisfies physical laws.}
    \label{fig:problem}
    %\vskip -0.1in
\end{figure}
We propose an Energy-consistent Neural Operator (ENO) as a general framework for learning solution operators of the systems that follow the energy conservation or dissipation law.
Our basic idea is to introduce physics-informed inductive bias into the operator learning framework to bias the output of the DNN-based solution operator that satisfies the laws of physics (see Fig.~\ref{fig:problem}).
\color{\mycolor}
ENO is based on the energy-based theory and simultaneously estimate solution operators and gradient flows of energy functional from data.
This allows ENO to consider the energetic behavior of the system without explicit PDEs, unlike existing approaches (e.g., PI-DeepONet and PINO).
\color{black}

\color{\mycolor}
\paragraph{Problem Setting}
Suppose that we have the data (described in Section~\ref{sec:operator}) and the differential operator $\mathcal{G}$ (described in Section~\ref{sec:energy}); then our goal is to learn solution operators of Hamiltonian or dissipative PDEs.
\color{black}

\paragraph{Method}
\begin{figure}[t]
    \centering
    \includegraphics[width=120mm]{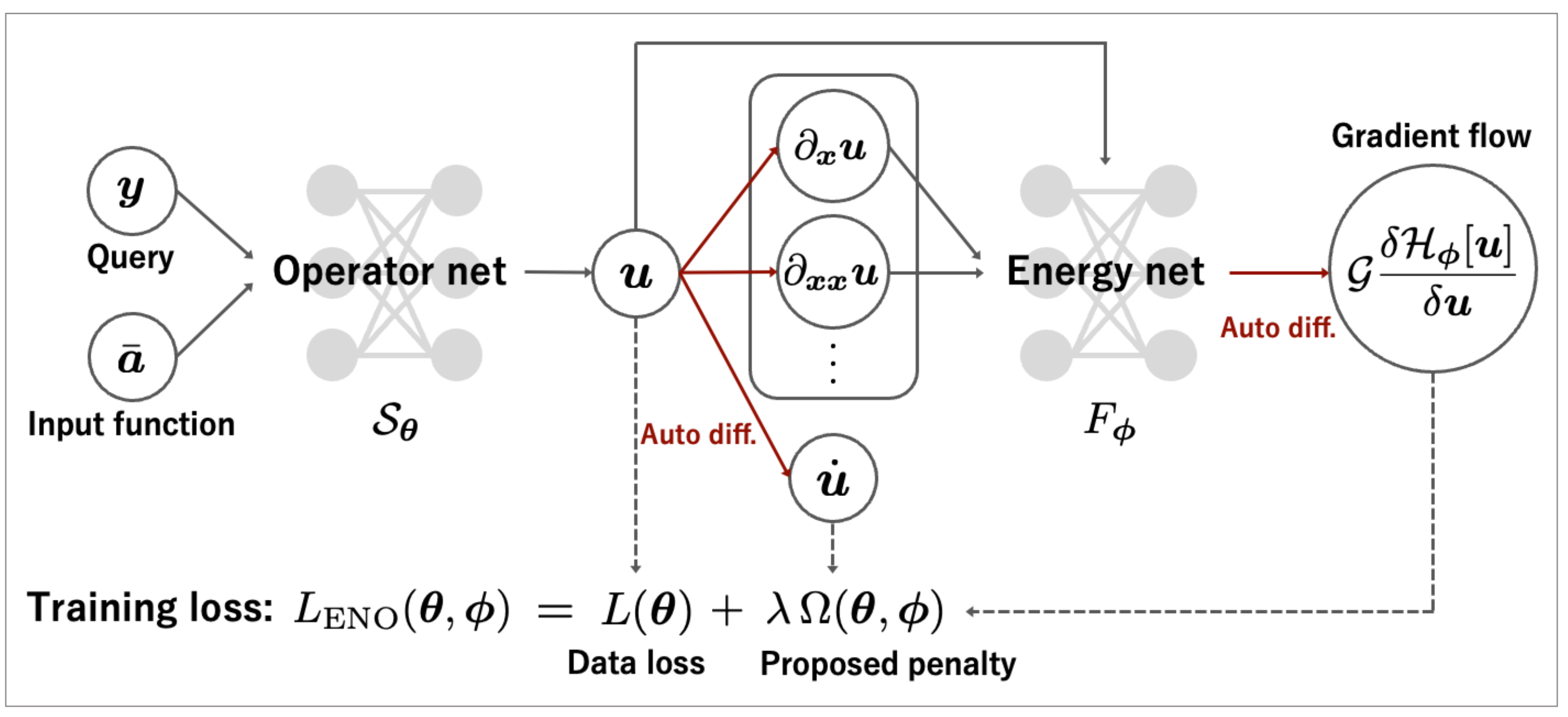}
    \caption{Schematic sketch of the architecture and training loss of ENO.
    Red arrows indicate automatic differentiation.
    ENO contains two networks: {\em operator net} and {\em energy net} parameterize solution operator and energy functional, respectively.
    By simultaneously minimizing data loss $L(\bm{\theta})$ and penalty $\Omega(\bm{\theta},\bm{\phi})$ inspired by energy-based theory, we can obtain a solution operator (i.e., operator net) to predict a solution that follows energy conservation or dissipation law without explicit PDEs.
    }
    \label{fig:eno}
\end{figure}
Fig.~\ref{fig:eno} schematically shows an architecture and a training loss of ENO.
\color{\mycolor}
We model the solution operator and the gradient flow using DNNs.
\color{black}
We employ the existing DNN models with parameters $\bm{\theta}$ for approximating the solution operator $\mathcal{S}$ as $\mathcal{S}_{\bm{\theta}}$, which we call {\em operator net} (see the left of Fig.~\ref{fig:eno}).
Note that any DNN architecture can be used as the operator net, such as the multi-layer perceptron (MLP), as long as it is differentiable; one can also use the advanced architectures (e.g., DeepONet and FNO) in our framework.

\color{\mycolor}
To model the gradient flow~\eqref{eq:H_pde}, we first parameterize energy functional $\mathcal{H}$~\eqref{eq:energy} as follows:
\begin{align}
    \mathcal{H}_{\bm{\phi}}[\bm{u}^{\bm{\theta}}]
        &= \int_{\mathcal{X}} F_{\bm{\phi}}\left(\bm{u}^{\bm{\theta}},\partial_{\bm{x}}\bm{u}^{\bm{\theta}},\partial_{\bm{xx}}\bm{u}^{\bm{\theta}},\ldots
        \right) \,d\bm{x},
        \label{eq:param_energy}
\end{align}
where we model the energy density $F$ as $F_{\bm{\phi}}$ using another DNN with parameters $\bm{\phi}$, which we call {\em energy net} (see the right of Fig.~\ref{fig:eno}). 
The order of partial differentiation in~\eqref{eq:param_energy} is mentioned in the {\bf Hyperparameters} paragraph.
Note that $\bm{u}^{\bm{\theta}}$ in~\eqref{eq:param_energy} represents that $\bm{u}$ is parameterized by $\bm{\theta}$, as $\bm{u}^{\bm{\theta}}(\bm{y})=\mathcal{S}_{\bm{\theta}}[\bar{\bm{a}}](\bm{y})$.
We can obtain the gradient flow of energy functional by calculating the functional derivative of $\mathcal{H}_{\bm{\phi}}$~\eqref{eq:functional} via automatic differentiation as follows:
\begin{align}
    \frac{\delta\mathcal{H}_{\bm{\phi}}[\bm{u}^{\bm{\theta}}]}{\delta u_m^{\bm{\theta}}}
        &= \frac{\partial F_{\bm{\phi}}}{\partial u_m^{\bm{\theta}}}
        -\sum_{d=1}^D 
        \left\{
            \frac{\partial}{\partial x_d}
            \left(
                \frac{\partial F_{\bm{\phi}}}{\partial u_{m,d}^{\bm{\theta}}}
            \right)
        \right\}
        +\cdots,
    \label{eq:param_gradient}
\end{align}
for $m=1,\ldots,M$.
Notice that one can use any DNN architecture as the energy net, such as the MLP, as long as it is differentiable.
\color{black}

The loss function of ENO is defined by
\begin{align}
    L_{\rm{ENO}}(\bm{\theta},\bm{\phi})
    = L(\bm{\theta}) + \lambda\, \Omega(\bm{\theta},\bm{\phi}),
    \label{eq:eno_loss}
\end{align}
where $L(\bm{\theta})$ is the data loss~\eqref{eq:data_loss} of the standard operator learning, $\Omega(\bm{\theta},\bm{\phi})$ is the proposed penalty, and $\lambda\in\mathbb{R}_{\geq 0}$ is a hyperparameter.
The penalty $\Omega(\bm{\theta},\bm{\phi})$ in~\eqref{eq:eno_loss} is given by
\begin{align}
    \Omega(\bm{\theta},\bm{\phi})
    = \frac{1}{I} \sum_{i=1}^I 
        \left(
            \frac{1}{K} \sum_{k=1}^{K}
            \biggl\| 
                %\dot{\bm{u}}_{\bm{\theta}}[\bm{a}_i](\bm{y}_{k})
                \dot{\bm{u}}^{\bm{\theta}}_i(\bm{y}_{k})
                %- \mathcal{G}\frac{\delta \mathcal{H}[\bm{u}_{\bm{\theta}}]} {\delta \bm{u}},
                - \mathcal{G}\frac{\delta \mathcal{H}_{\bm{\phi}}[\bm{u}^{\bm{\theta}}_i]}{\delta \bm{u}^{\bm{\theta}}_i}(\bm{y}_{k})
            \biggr\|^2
    \right),
    \label{eq:omega}
\end{align}
where $\bm{u}^{\bm{\theta}}_i=\mathcal{S}_{\bm{\theta}}[\bar{\bm{a}}_i]$ and $\dot{\bm{u}}^{\bm{\theta}}_i$ denotes its time-derivative that can be obtained by utilizing automatic differentiation.
\color{\mycolor}
Our penalty function~\eqref{eq:omega} can be considered at $K\in\mathbb{N}$ arbitrary query points $\{\bm{y}_{k}\mid k=1,\ldots,K\}\subset\mathcal{Y}$, potentially not included in training data points $\bar{\mathcal{Y}}_1\cup\cdots\cup\bar{\mathcal{Y}}_I$.
\color{black}
By considering the penalty, we can introduce the inductive bias to the time-derivative $\dot{\bm{u}}^{\bm{\theta}}_i$ of the solution to ensure the energy conservation or dissipation law.

Parameters $\bm{\theta}$ (for operator net) and $\bm{\phi}$ (for energy net) are estimated by minimizing the loss~\eqref{eq:eno_loss}; ENO can infer not only the solution operator but also the energy functional from observed data.
The training procedure for ENO is shown in Algorithm~\ref{alg}.
In line 6 of Algorithm~\ref{alg}, we uniformly sample $K$ query points $\{\bm{y}_k\}$ and add up the penalty terms at the sampled query points.
%This scheme resembles the concept of {\em collocation points} in physics-informed neural networks~\cite{raissi:physics}.
This scheme yields the smoothing effect based on the laws of physics over the entire spatio-temporal domain, allowing the appropriate training of solution operators, even when training data resolution is lower, such as super-resolution settings.

ENO can be used for Hamiltonian or dissipative systems described by ODEs or PDEs, and is widely applicable to various physical phenomena covered by the energy-based theory (described in Section~\ref{sec:energy}).

\begin{algorithm}[t]
    \caption{Training procedure for ENO}
    \label{alg}
\begin{algorithmic}[1]
    \STATE {\bfseries Input:} Data $\{(\bar{\bm{a}}_i, \bar{\bm{u}}_i)\}_{i=1}^I$, data points $\mathcal{Y}=\bar{\mathcal{Y}}_1\cup\cdots\cup\bar{\mathcal{Y}}_I$, differential operator $\mathcal{G}$, mini-batch size $I_{\rm{b}}$, number of query points $K$ for penalty, hyperparameter $\lambda$
    \STATE {\bfseries Output:} Trained DNN parameters $\bm{\theta}$, $\bm{\phi}$
    \STATE Initialize DNN parameters $\bm{\theta}$, $\bm{\phi}$.
    \REPEAT
    \STATE Randomly sample $I_{\rm{b}}$ indices from $\{1,\ldots,I\}$.
    \STATE Uniformly sample $K$ query points $\{\bm{y}_k\}$ for penalty, all of which are contained in domain $\mathcal{Y}$.
    \color{\mycolor}
    \STATE /* Predict solutions */
    \STATE Predict $\bm{u}$ at all points $\mathcal{Y}\cup\{\bm{y}_k\}$ for respective input functions $\{\bar{\bm{a}}_i\}$ via operator net $\mathcal{S}_{\bm{\theta}}$.
    \STATE /* Estimate gradient flows  */
    \STATE Obtain partial derivatives $\dot{\bm{u}}, \partial_{\bm{x}}\bm{u}, \partial_{\bm{xx}}\bm{u},\ldots$ at sampled query points $\{\bm{y}_k\}$ via automatic differentiation.
    \STATE Calculate energy~\eqref{eq:param_energy} using estimated solutions and partial derivatives via energy net $F_{\bm{\phi}}$.
    \STATE Obtain flows~\eqref{eq:param_gradient} via automatic differentiation.
    \STATE /* Update parameters */
    \color{black}
    %\STATE Calculate the ENO loss~\eqref{eq:eno_loss}.
    \STATE Update DNN parameters $\bm{\theta}$, $\bm{\phi}$ using the gradient of the ENO loss~\eqref{eq:eno_loss} via a stochastic gradient method.
    \UNTIL{End condition is satisfied.}
\end{algorithmic}
\end{algorithm}

\paragraph{Hyperparameters}
ENO has two hyperparameters to be determined.
The first is $\lambda$ in~\eqref{eq:eno_loss} that controls the penalty for violating physical constraints.
\color{\mycolor}
The second is the order of partial differentiation in~\eqref{eq:param_energy}.
These hyperparameters can be determined based on the validation error.
\color{black}

\section{Experiments}
\label{sec:experiments}
We demonstrate the effectiveness of our proposal, ENO, using simulation data of ODE and PDE systems.
This section provides the experiments on Hamiltonian and dissipative PDEs.
The experiments on ODE systems are described in Appendix~\ref{app:ode}.
We focus on tasks that predict solution functions when we are given initial conditions as input functions.

\paragraph{Data}
We generated simulation data of PDE systems whose energy functional $\mathcal{H}$ is known.
Notice that the explicit form of $\mathcal{H}$ was used only for data generation and {\em not for training}.
We evaluated the proposed ENO on a Hamiltonian PDE, namely the one-dimensional Korteweg--de Vries (KdV) equation~\cite{korteweg} under the periodic boundary condition, which is a shallow water wave equation defined on the time-space $\mathcal{T}\times\mathcal{X}=[0,0.5]\times[0,10]$.
The energy functional $\mathcal{H}$ for the function $u:\mathcal{T}\times\mathcal{X}\to\mathbb{R}$ is given by $\mathcal{H}[u]=\int_{\mathcal{X}} F(u,\partial_{x}u)\,dx$, where
\begin{align}
F(u,\partial_{x}u)=u^3-\frac{1}{2}(\partial_x u)^2.
\end{align}
An input function (i.e., an initial condition) $a:\mathcal{X}\to\mathbb{R}$ was set to a sum of two solitons, represented by 
\begin{align}
a(x) = \sum_{i=1}^2 2\kappa_i^2{\rm sech}^2\left(\kappa_i(x-d_i)\right), 
\end{align}
where we set $d_1=3$ and $d_2=6$, and $\kappa_1$ and $\kappa_2$ were uniformly sampled across ranges $[0.5,1.0]$ and $[1.5,2.0]$, respectively.

We also evaluated ENO on a dissipative PDE, namely the one-dimensional Cahn–Hilliard equation~\cite{cahn} under the periodic boundary condition, which is often used for modeling a phase separation of copolymer melts, defined on the time-space $\mathcal{T}\times\mathcal{X}=[0,0.05]\times[0,1]$.
The energy functional $\mathcal{H}$ for the function $u:\mathcal{T}\times\mathcal{X}\to\mathbb{R}$ is given by $\mathcal{H}[u]=\int_{\mathcal{X}} F(u,\partial_x u)\,dx$, where
\begin{align}
    F(u,\partial_x u)=\frac{1}{4}u^4 - \frac{1}{2}u^2 + \frac{\gamma}{2}(\partial_x u)^2.
\end{align}
Here, the coefficient $\gamma\in\mathbb{R}_{>0}$ denotes the mobility of the monomers, which we set to 0.0005.
We used the orthogonal polynomials of degree five as an initial condition $a:\mathcal{X}\to\mathbb{R}$, represented by
\begin{align}
    a(x) = \sum_{i=1}^5 \beta_i C_i(x),
\end{align}
where $C_i(x)$ are Chebyshev polynomials of the first kind, and $\beta_i$ were uniformly sampled across a range $[0,0.05]$.

To generate trajectory data of $u$ from respective initial conditions, we first discretize PDEs~\eqref{eq:H_pde} in the spatial domain, where we used an appropriate discretization introduced in~\cite{celledoni:preserving} to ensure the energy conservation or dissipation law.
Then, we obtained trajectories by applying a numerical solver, i.e., the Dormand--Prince method with adaptive time-stepping (implemented in SciPy), to the discretized PDEs.
Here, the relative and absolute tolerances were set to $10^{-12}$ and $10^{-14}$, respectively.
Space was uniformly discretized to $N_{\rm x}=100$ cells in $\mathcal{X}$, and time was uniformly discretized to $N_{\rm t}=1000$ points in $\mathcal{T}$.

\paragraph{Task}
In our experiments, we considered a super-resolution setting; we predicted the high-resolution test data from the low-resolution training data. 
We first generated 1000 trajectories from different initial conditions with high-resolution: $(N_{\rm x}, N_{\rm t})=(100,1000)$, of which 90\% were used for training and 10\% for validation.
We then created three different resolutions $(N_{\rm x},N_{\rm t})\in\{(10,10),(15,15),(20,20)\}$ of data by downsampling the original high-resolution data.
We generated 100 test trajectories with high-resolution, whose initial conditions are different from the training and validation data.
The experiments were conducted five times by resampling the training and validation sets.

\begin{table}[!t]
%\vskip -0.03in
\caption{Average MSEs for PDEs when using three different resolutions of training data.
The best results are emphasized by bold font.
}
\label{tb:result_pde}
%\vskip -0.25in
\begin{center}
\begin{small}
\begin{sc}
\subtable[KdV equation: Multiplied by $10^4$ for \textsc{Traj.} and by $10^{2}$ for \textsc{Mass}.]{
%\scalebox{0.75}{
\begin{tabular}{rrrr|rrr|rrr} \toprule
    &\multicolumn{3}{c|}{$(N_{\rm{x}},N_{\rm{t}})=(10,10)$} 
    &\multicolumn{3}{c|}{$(N_{\rm{x}},N_{\rm{t}})=(15,15)$}
    &\multicolumn{3}{c}{$(N_{\rm{x}},N_{\rm{t}})=(20,20)$}
    \\
    \cmidrule(l){2-4} 
    \cmidrule(l){5-7} 
    \cmidrule(l){8-10}
    &Traj. &Energy &Mass 
    &Traj. &Energy &Mass 
    &Traj. &Energy &Mass\\ \midrule
    ENO 
    &{\bf 2.14} &{\bf 3.38} &{\bf 3.88}
    &{\bf 0.82} &{\bf 0.34} &{\bf 1.49}
    &{\bf 0.53} &{\bf 0.12} &0.75
    %&0.57 &0.15 &0.74
    \\[1pt]
    ENO (fixed) 
    &31.07 &36.16 &83.16
    &1.05 &0.45 &2.42 
    &0.54 &0.14 &{\bf 0.62}
    \\[1pt]
    Vanilla NO 
    &77.23 &132.13 &187.59
    &26.55 &25.85 &18.85
    &0.69 &0.15 &0.63
    %&0.68 &0.13 &0.65
    \\[1pt]
    DeepONet 
    &91.82 &168.18 &359.8
    &47.54 &42.70 &80.35
    &5.22 &1.45 &7.86
    \\[1pt]
    \bottomrule
\end{tabular}
%}
}
%\vskip -0.01in
\subtable[Cahn–Hilliard equation: Multiplied by $10^3$ for \textsc{Traj.}, by $10^{6}$ for \textsc{Energy}, and by $10^{1}$ for \textsc{Mass}.]{
%\scalebox{0.75}{
\begin{tabular}{rrrr|rrr|rrr} \toprule
    &\multicolumn{3}{c|}{$(N_{\rm{x}},N_{\rm{t}})=(10,10)$} 
    &\multicolumn{3}{c|}{$(N_{\rm{x}},N_{\rm{t}})=(15,15)$}
    &\multicolumn{3}{c}{$(N_{\rm{x}},N_{\rm{t}})=(20,20)$}
    \\
    \cmidrule(l){2-4} 
    \cmidrule(l){5-7} 
    \cmidrule(l){8-10}
    &Traj. &Energy &Mass 
    &Traj. &Energy &Mass 
    &Traj. &Energy &Mass\\ \midrule
    ENO 
    &{\bf 70.88} &{\bf 62.43} &{\bf 9.98}
    &{\bf 2.71} &{\bf 16.42} &{\bf 2.76}
    &{\bf 0.45} &{\bf 0.71} &{\bf 0.22}
    %&0.16 &0.22 &0.89
    \\[1pt]
    ENO (fixed)
    &154.13 &1461.30 &137.83
    &12.49 &127.55 &5.44
    &0.77 &1.59 &0.29 
    \\[1pt]
    Vanilla NO 
    &193.32 &1918.18 &436.69
    &39.41 &1783.07 &29.95
    &0.87 &1.86 &0.30
    %&0.15 &0.21 &1.55
    \\[1pt]
    DeepONet 
    &127.76 &895.19 &301.72
    &18.54 &425.36 &11.81
    &3.39 &13.91 &0.94
    \\[1pt]
    \bottomrule
\end{tabular}
%}
}
\end{sc}
\end{small}
\end{center}
%\vskip -0.25in
\end{table}

\begin{figure}[tb]
%\vskip 0.2in
\begin{center}
\subfigure[Results for KdV equation ($(N_{\rm x}, N_{\rm t})=(10,10)$).]{\includegraphics[height=50mm]{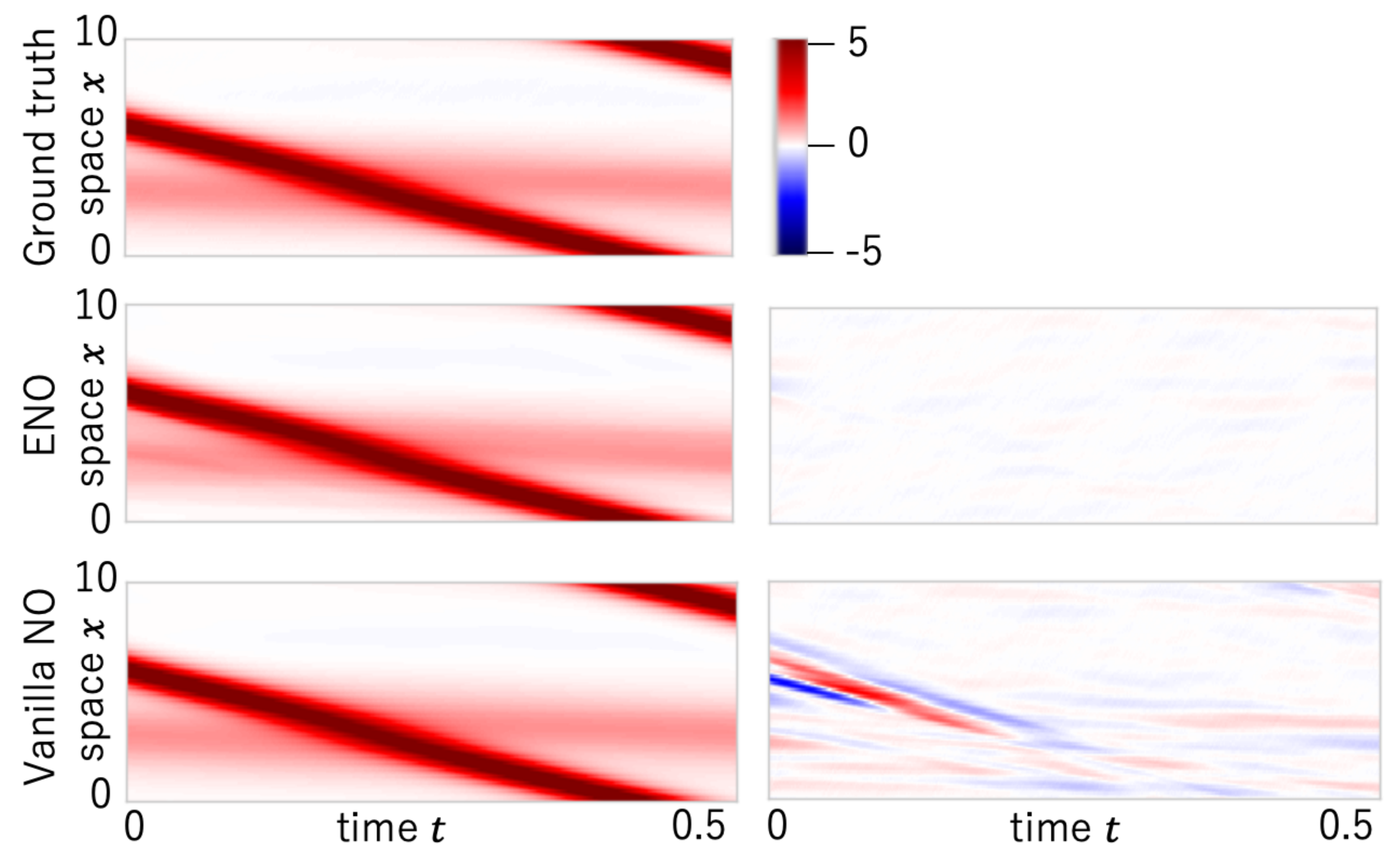}\label{fig:}}
%\hspace{5mm}
\subfigure[Results for Cahn-Hilliard equation ($(N_{\rm x}, N_{\rm t})=(15,15)$).]{\includegraphics[height=50mm]{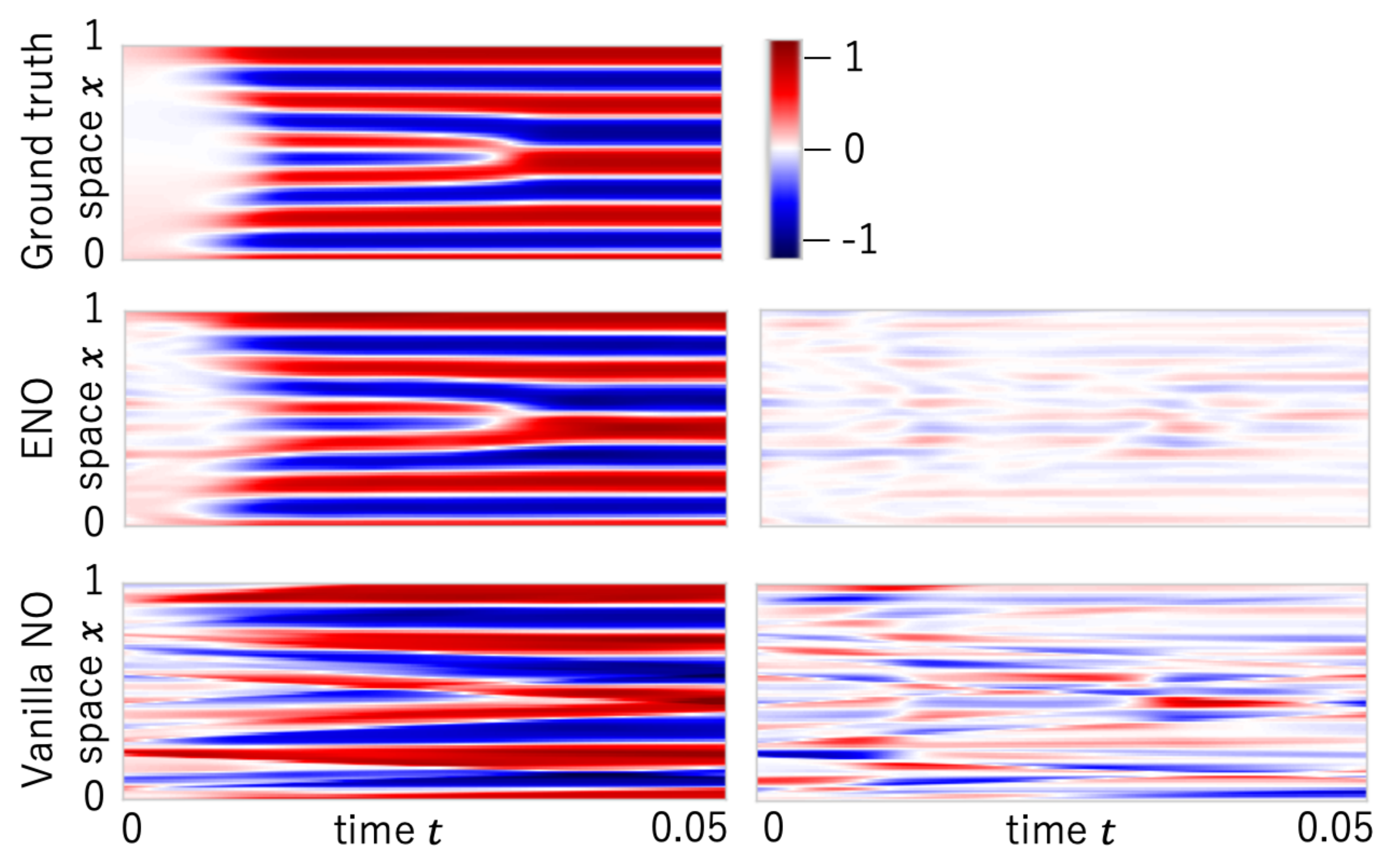}\label{fig:}}
\caption{Visualization of the predicted solutions.
Right column for each system is the difference between ground truth and its prediction, where the difference values for KdV equation were multiplied by 4.}
\label{fig:pde_vis}
\end{center}
%\vskip -0.2in
\end{figure}

\paragraph{ENO Setup}
%\noindent {\textbf{ENO Setup}.}
We adopted two multi-layer perceptrons (MLPs) to implement the operator net $\mathcal{S}_{\bm{\theta}}[a](\bm{y})={\rm MLP}(\bar{a}\oplus\bm{y})$ and the energy net $F_{\bm{\phi}}(u,\partial_xu)={\rm MLP}(u\oplus\partial_xu)$, where $\oplus$ denotes the concatenation operator, and the order of partial differentiation was assumed to be known.
The respective nets had three layers, 200 hidden units, and tanh activations.
We trained them by minimizing the ENO loss~\eqref{eq:eno_loss}, where the validation data were used for early stopping, and the maximum number of epochs was 10000.
We used the Adam optimizer~\cite{kingma:adam} implemented in PyTorch~\cite{pytorch}, and set the learning rates for $\bm{\theta}$ and $\bm{\phi}$ to $10^{-3}$ and $10^{-4}$, respectively.
Mini-batch size $I_{{\rm b}}$ in Algorithm~\ref{alg} was 30.
Query points $\{\bm{y}_k\}_{k=1}^K$ for the penalty~\eqref{eq:omega} were uniformly sampled in the spatio-temporal domain, where $K$ was set to 200.
\color{\mycolor}
Hyperparameter $\lambda$ in~\eqref{eq:eno_loss} was chosen from $\{10^{-8},10^{-7},\cdots,10^{-1}\}$ based on the loss~\eqref{eq:data_loss} for the validation data.
\color{black}

\color{\mycolor}
One benefit of ENO is that it can consider the penalty term~\eqref{eq:omega} at uniformly sampled query points in Algorithm~\ref{alg}. 
To verify its effectiveness, we prepared a variant (called \textsc{ENO (fixed)}) considering the penalty term evaluated only at fixed data points in training data.
If fewer than 200 points were available in training data, all points were used as queries; otherwise, 200 points were randomly selected from fixed training data points at each epoch.
\color{black}

\paragraph{Baselines}
To evaluate the effectiveness of the proposed penalty~\eqref{eq:omega}, we compared ENO with a vanilla neural operator (called \textsc{Vanilla NO}) implemented by the MLP, which corresponds to the method that excludes the penalty term from our ENO.
We also adopted one of the most widely-used method that can learn solution operators from data (without explicit PDEs), namely the deep operator network (\textsc{DeepONet})~\cite{lu2021learning}, as the baseline.
In \textsc{DeepONet}, the solution operator is modeled by an inner product of two latent variables $\bm{z}(\bar{\bm{a}}), \bm{z}(\bm{y}) \in \mathbb{R}^Q$, as $\mathcal{S}_{\bm{\theta}}[\bar{\bm{a}}](\bm{y})=\bm{z}(\bar{\bm{a}})^\top \bm{z}(\bm{y})$, where $\bm{z}(\bar{\bm{a}})$ (called branch net) and $\bm{z}(\bm{y})$ (called trunk net) are modeled by any neural network with $\bar{\bm{a}}$ and $\bm{y}$ as inputs, respectively.
In our experiments, the branch and trunk nets were modeled by an MLP with three layers, 200 hidden units, and tanh activations.
Dimension $Q$ of the latent variables was set to 30.
The other settings were identical to those of ENO.

%Although another typical method is the Fourier neural operator (FNO)~\cite{li2021fourier}, its standard implementation is for the prediction of $\bm{u}$ at discrete time steps (not at a continuous timeline) predefined at the training phase; hence, it cannot be straightforwardly applied to the super-resolution task in the spatio-temporal domain described in the {\bf Task} paragraph; we did not adopt it as a baseline.

\paragraph{Results}
Table~\ref{tb:result_pde} shows the mean squared error (MSE) between the true and predicted solution trajectories for ENO and the baselines (see the culumn \textsc{Traj.}).
It also shows the MSE for energy and mass calculated using the predicted trajectories (see the culumns \textsc{Energy} and \textsc{Mass}).
\color{\mycolor}
Here, we provided an average of MSEs over five trials; we omitted the standard deviations for readability (see Appendix~\ref{app:pde} for the full results).
\color{black}
In all cases, ENO achieved comparable or better performance than the baselines (i.e., \textsc{Vanilla NO} and \textsc{DeepONet}) regarding trajectory, energy, and mass; the performance improvements were significant in the settings where the training data resolution was lower.
These results indicate that our ENO can accurately predict solutions while capturing
\color{\mycolor}
the energy conservation or dissipation law, even when only the lower-resolution data is available for training. 
The energetic behavior estimated by each method is shown in Appendix~\ref{app:pde}.
In addition, the errors of ENO were lower than those of \textsc{ENO (Fixed)}, especially in the lower-resolution settings. 
This indicates the effectiveness of considering the penalty terms at uniformly sampled query points (see Algorithm~\ref{alg}) in such settings. 
\color{black}

Fig.~\ref{fig:pde_vis} visualizes the solutions predicted by ENO and \textsc{Vanilla NO}. 
%when the training data resolution was $(N_{\rm{x}},N_{\rm{t}})=(15,15)$.
Visualization results with other methods are shown in Appendix~\ref{app:pde}.
As shown in Fig.~\ref{fig:pde_vis}, our ENO more appropriately captured the physical behavior than \textsc{Vanilla NO}.
This result indicates that our primary contribution, the penalty function inspired by the energy-based theory, can significantly improve the predictive performance of naive DNN-based operators.

\paragraph{Computational Time}
The average training time of ENO was 1.46 hours and 3.77 hours for the KdV equation and Cahn-Hilliard equation, respectively, when the data resolution was $(N_{\rm{x}},N_{\rm{t}})=(15,15)$.
In testing, ENO took only 0.15 seconds to obtain one solution. 
The experiments were conducted on a single NVIDIA A100 GPU.

\section{Conclusion}
\label{sec:conclusion}
We proposed an Energy-consistent Neural Operator (ENO) to train neural network-based solution operators, which can predict physical behaviors that adhere to the energy conservation or dissipation law without explicit PDEs.
Our significant contribution is a penalty function derived from the energy-based theory, which is a general tool that has the potential to be applied to various operator learning problems.
We experimentally confirmed ENO's effectiveness using ODE and PDE systems.

\paragraph{Limitation}
\color{\mycolor}
In this study, we assume that the differential operator $\mathcal{G}$ in~\eqref{eq:omega} is known. 
Our future work is to extend ENO to be able to estimate $\mathcal{G}$ from data.
Another limitation is the computational cost. The order of partial differentiation input to energy net $F_{\bm{\phi}}$ in~\eqref{eq:param_energy} might be higher for large-scale and complex systems. 
In that case, the computational cost involved in automatic differentiation becomes prohibitive. 
It is desirable to develop an efficient implementation method for the ENO framework.
\color{black}

%Bibliography
\bibliographystyle{unsrt}  
%\bibliography{reference}  

\appendix
\section{Experiments on ODE systems}
\label{app:ode}
In this section, we demonstrate the effectiveness of our proposed method, ENO, using the simulation data of several Hamiltonian systems.
In the setting of the ODE systems, although the initial condition is a constant, not a function, it can be regarded as a special case of operator learning.

\paragraph{Data}
We evaluated the ENO using three Hamiltonian systems: a mass-spring, a pendulum, and a Duffing oscillator.
The system's state is defined by $\bm{u}=(q,p)$, where $q:\mathcal{T}\to\mathbb{R}$ and $p:\mathcal{T}\to\mathbb{R}$ are the generalized coordinate and the generalized momentum, respectively.
The Hamiltonian (i.e., energy function) of the mass-spring is
\begin{align}
\mathcal{H} = \frac{1}{2}kq^2 + \frac{p^2}{2m},
\end{align}
where $k$ is the spring constant and $m$ is the mass constant.
In the experiments, we set $k=m=1$.
The Hamiltonian of the pendulum is
\begin{align}
    \mathcal{H} = mgl(1-\cos{q}) + \frac{p^2}{2ml^2},
\end{align}
where we denote the gravitational constant by $g$ and the pendulum's length by $l$.
In the experiments, we set $g=3$ and $m=l=1$.
The Hamiltonian of the Duffing oscillator is
\begin{align}
    \mathcal{H} 
    =\frac{1}{2}p^2 + \frac{\alpha}{2} q^2 + \frac{\beta}{4}q^4,
\end{align}
where we set parameters $\alpha=\beta=1$.

We sampled initial conditions $\bm{a}=(q,p)$, where $q$ was uniformly distributed across a predefined range, and $p$ was fixed to zero; the range of $q$ for the mass-spring and the pendulum was $[1.3,2.3]$, and the range for the Duffing oscillator was $[1.7,2.0]$.
The state trajectories were generated by employing a numerical integrator, i.e., the Dormand--Prince method with adaptive time-stepping, implemented in SciPy.
We evaluated the effectiveness of ENO against the data resolution by preparing 100 trajectories sampled at frequencies of 2 and 10 Hz.
The observation period was 10 seconds for the mass-spring and the Duffing oscillator and 5 seconds for the pendulum.
We randomly split the data and used 80\% for training and 20\% for validation.
We independently generated a test set of 100 trajectories from the training and validation sets.
The frequency of the test trajectory was 100 Hz, and the observation period was identical to the training and validation sets.
The experiments were conducted five times by resampling the training and validation sets.

\paragraph{ENO Setup}
For implementation, we used the MLP that had five layers, 32 hidden units, and tanh activations.
Mini-batch size $I_{{\rm b}}$ in Algorithm~\ref{alg} was 20.
Query points $\{\bm{y}_k\}_{k=1}^K$ for the penaly~\eqref{eq:omega} were uniformly sampled in the time domain, where $K$ was set to 20.
Hyperparameter $\lambda$ in~\eqref{eq:eno_loss} was chosen from $\{0.001,0.01,0.1,0.5,1.0,2.0\}$ based on the loss~\eqref{eq:data_loss} for the validation data.
The other settings were identical as described in Section~\ref{sec:experiments}.

\paragraph{Baselines}
We compared the ENO with the baselines: \textsc{Vanilla NO} and \textsc{DeepONet}.
For all methods, the MLP used for implementation had five layers, 32 hidden units, and tanh activations.
In \textsc{DeepONet}, dimension $Q$ of the latent variables was set to ten.
The other settings were identical as described in Section~\ref{sec:experiments}.

\begin{table}[!t]
 %\small
 \caption{Average MSEs for ODEs when frequency was 2 or 10 Hz: bold font indicates statistically significant differences between our method and baselines (a paired t-test) at level of $P<0.05$.
 Values in parentheses represent improvement rates of our method over baselines.
 Red letters indicate improvement rates of 50\% or more.
 }
 %\vspace{-12pt}
 \label{tb:MSE_ode}
 \begin{center}
 \begin{small}
 \begin{sc}
 \subtable[Mass-spring: all values are multiplied by $10^4$.]
 {
  \begin{tabular}{r r r r r r r} 
  \toprule
  &\multicolumn{2}{c}{ENO} 
  &\multicolumn{2}{c}{Vanilla NO} 
  &\multicolumn{2}{c}{DeepONet}
  \\
  \cmidrule(l){2-3} 
  \cmidrule(l){4-5} 
  \cmidrule(l){6-7}  
  Frequency &Traj. &Energy &Traj. &Energy & Traj. &Energy
  \\ \midrule
   2 
   &{\bf 2.61} (23\%) &9.79 (11\%) 
   &3.41 &10.95
   &1.46 &4.93 
   \\[1pt]
   10 
   &{\bf 2.50} (28\%) &9.12 (6\%) 
   &3.50 &9.73 
   &1.37 &4.31 
   \\[1pt]
   \bottomrule
  \end{tabular}
  }
 \subtable[Pendulum: all values are multiplied by $10^3$.]{
  \begin{tabular}{r r r r r r r} 
  \toprule
  &\multicolumn{2}{c}{ENO} 
  &\multicolumn{2}{c}{Vanilla NO} 
  &\multicolumn{2}{c}{DeepONet} 
  \\
  \cmidrule(l){2-3} 
  \cmidrule(l){4-5} 
  \cmidrule(l){6-7} 
  Frequency &Traj. &Energy &Traj. &Energy & Traj. &Energy
  \\ \midrule
  2 
  &{\bf 0.52} ({\color{red}{64\%}}) & {\bf 1.75} ({\color{red}{76\%}}) 
  &1.43 &7.25 
  &4.20 &32.35 
  \\[1pt]
  10 
  &{\bf 0.48} (27\%) &{\bf 1.55} ({\color{red}{52\%}}) 
  &0.67 &3.22 
  &1.62 &11.33 
  \\[1pt]
  \bottomrule
  \end{tabular}
  }
 \subtable[Duffing oscillator: all values are multiplied by $10^3$.]
 {
  \begin{tabular}{r r r r r r r} 
  \toprule
  &\multicolumn{2}{c}{ENO} 
  &\multicolumn{2}{c}{Vanilla NO} 
  &\multicolumn{2}{c}{DeepONet} 
  \\
  \cmidrule(l){2-3} 
  \cmidrule(l){4-5} 
  \cmidrule(l){6-7} 
  Frequency &Traj. &Energy &Traj. &Energy & Traj. &Energy
  \\ \midrule
  2 
  &{\bf 1.12} ({\color{red}{72\%}}) &{\bf 2.06} ({\color{red}{74\%}}) 
  &3.95 &7.97 
  &6.16 &13.26 
  \\[1pt]
  10 
  &{\bf 0.95} (33\%) &{\bf 2.11} ({\color{red}{55\%}}) 
  &1.43 &4.68 
  &4.61 &10.84 
  \\[1pt]
  \bottomrule
  \end{tabular}
  }
 \end{sc}
 \end{small}
 \end{center}
\end{table}

\paragraph{Results}
Table~\ref{tb:MSE_ode} shows the mean squared errors (MSEs) between the true and predicted trajectories for ENO and the baselines.
It also shows the MSEs of the energy $\mathcal{H}(\hat{\bm{u}})$ evaluated using predicted state $\hat{\bm{u}}$ and true energy $\mathcal{H}(\bm{u})$, where $\mathcal{H}(\cdot)$ is the true Hamiltonian of each system.
In all cases, the proposed method yielded lower errors than the baselines in terms of both trajectory and energy.
These results show that ENO can accurately predict physical dynamics while adhering to energy conservation law.
The values in parentheses in Table~\ref{tb:MSE_ode} represent the improvement rates of the proposed method over the baselines.
They were especially high at a sampling frequency of 2 Hz.
These results indicate that our proposed method is advantageous in settings of low-resolution data because, as discussed in Section~\ref{sec:eno},  our penalty~\eqref{eq:omega} addresses arbitrary query points not included in the training data.

\section{Additional results on PDE systems}
\label{app:pde}
Table~\ref{tb:result_pde_full} shows the mean squared error (MSE) and standard deviations between the true and predicted solution trajectories for ENO and the baselines (see the culumn \textsc{TRAJ.}). 
It also shows the MSEs and standard deviations for energy and mass calculated using the predicted trajectories (see the culumns \textsc{ENERGY} and \textsc{MASS}). 

\begin{figure}[t]
    \centering
    \includegraphics[width=160mm]{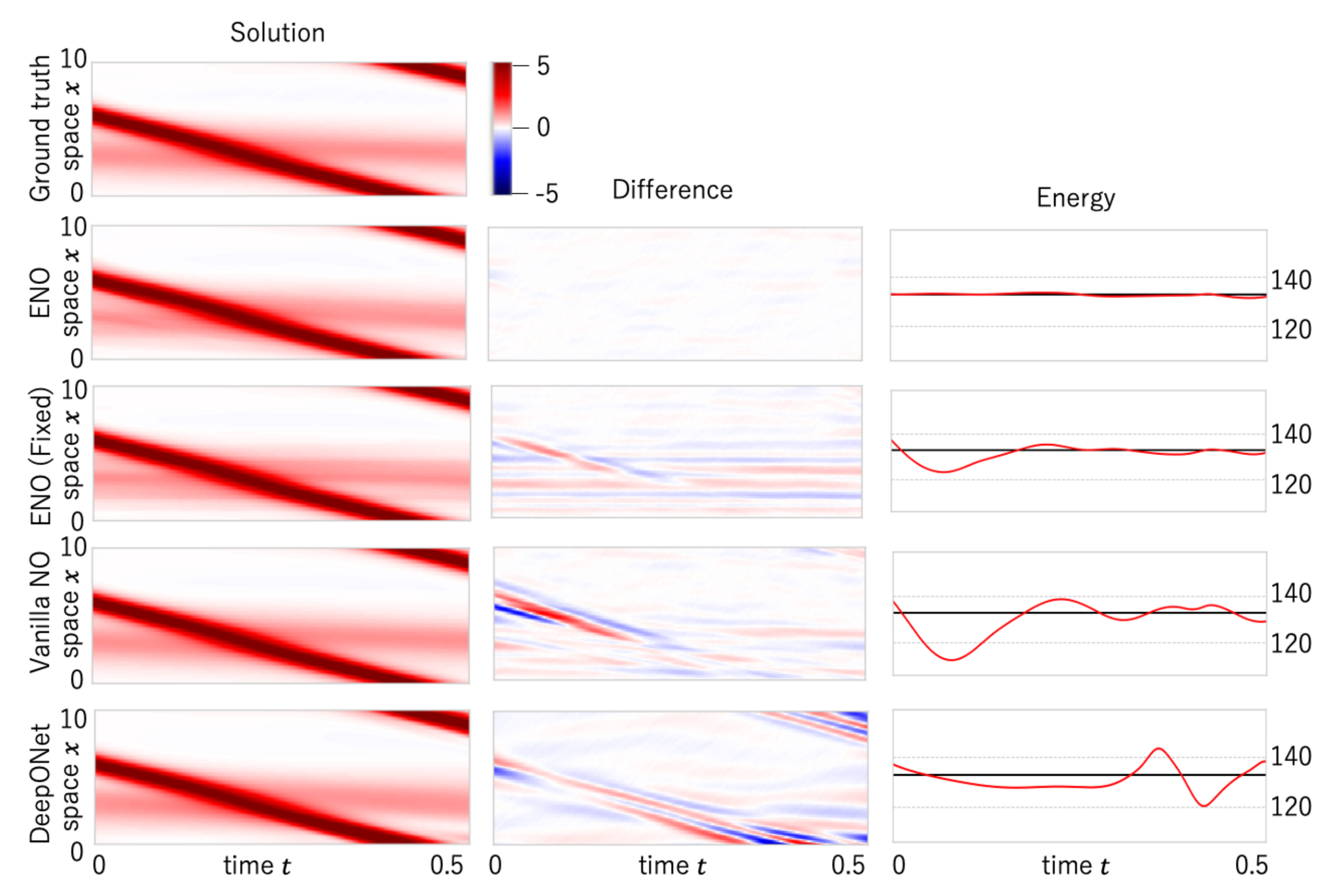}
    \caption{Results for KdV equation ($(N_{\rm x}, N_{\rm t})=(10,10)$).
    First column is a visualization of predicted solutions. 
    Second column is a difference between ground truth and its prediction, where the difference values were multiplied by 4. 
    Third column provides a comparison between the true energy (black line) and its estimate (red line).
    }
    \label{fig:kdv_vis_full}
\end{figure}

\begin{figure}[t]
    \centering
    \includegraphics[width=160mm]{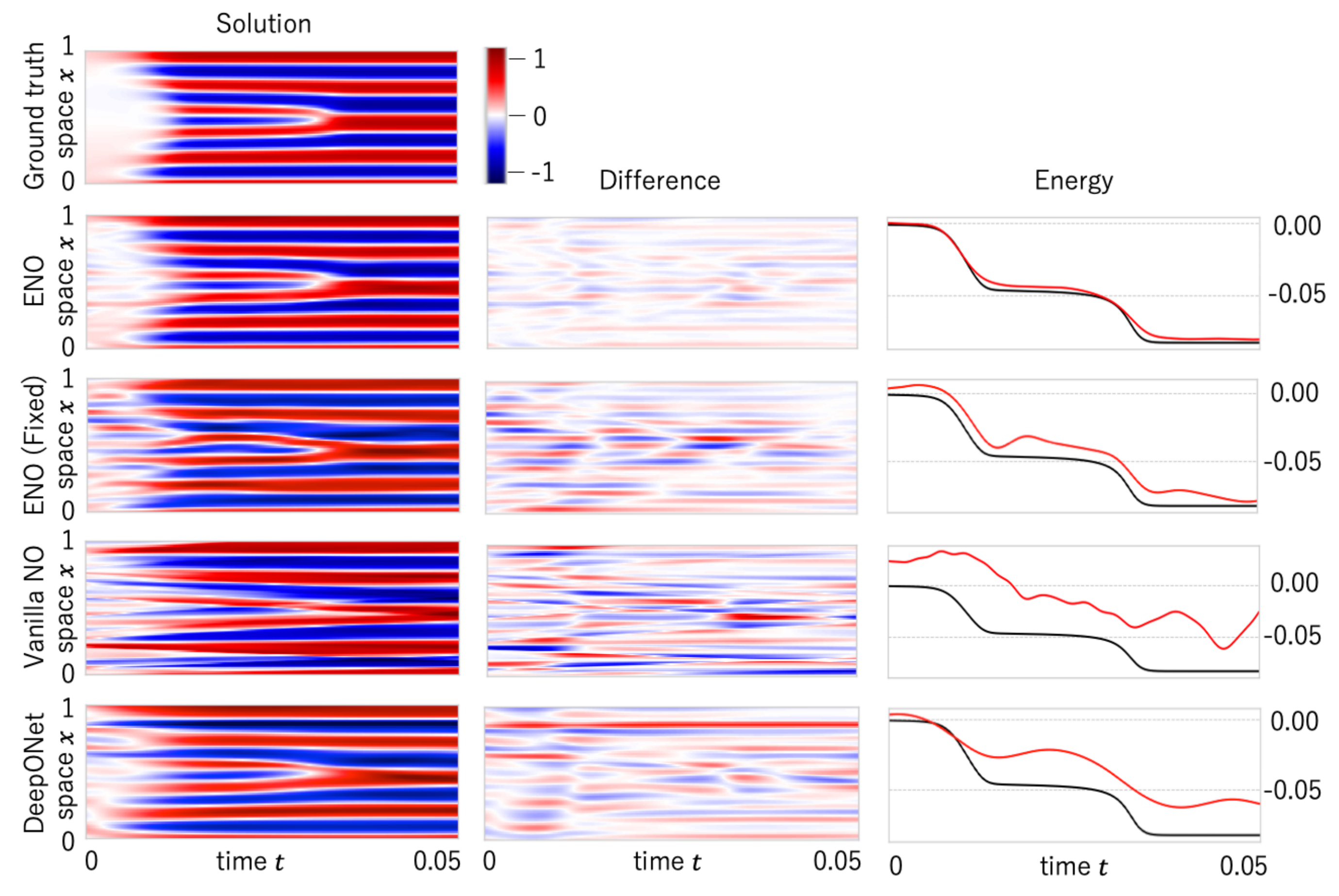}
    \caption{Results for Cahn-Hilliard equation ($(N_{\rm x}, N_{\rm t})=(15,15)$).
    First column is a visualization of predicted solutions. 
    Second column is a difference between ground truth and its prediction.
    Third column provides a comparison between the true energy (black line) and its estimate (red line).
    }
    \label{fig:ch_vis_full}
\end{figure}

Figures~\ref{fig:kdv_vis_full} and~\ref{fig:ch_vis_full} show the visualization results for the KdV equation and the Cahn-Hilliard equation, respectively.
As shown in the first and second columns of these figures, ENO can more accurately predict solutions than the other methods.
Moreover, the third columns of these figures show that ENO can capture the energy conservation (Fig.~\ref{fig:kdv_vis_full}) or dissipation law (Fig.~\ref{fig:ch_vis_full}) from data, without explicit PDEs.

\begin{landscape}
\begin{table}[!t]
\caption{MSEs and standard deviations for PDE systems when using three different resolutions of training data.
The best results are emphasized by bold font.
}
\label{tb:result_pde_full}
%\vskip -0.25in
\begin{center}
%\begin{small}
%\tiny
\begin{sc}
\subtable[KdV equation]{
\scalebox{0.68}{
\begin{tabular}{rrrr|rrr|rrr} \toprule
    &\multicolumn{3}{c|}{$(N_{\rm{x}},N_{\rm{t}})=(10,10)$} 
    &\multicolumn{3}{c|}{$(N_{\rm{x}},N_{\rm{t}})=(15,15)$}
    &\multicolumn{3}{c}{$(N_{\rm{x}},N_{\rm{t}})=(20,20)$}
    \\
    \cmidrule(l){2-4} 
    \cmidrule(l){5-7} 
    \cmidrule(l){8-10}
    &Traj. &Energy &Mass 
    &Traj. &Energy &Mass 
    &Traj. &Energy &Mass
    \\ \midrule
    ENO 
    &{\bf 2.14}$\pm$0.03 ($\times 10^{-4}$) &{\bf 3.38}$\pm$0.30 ($\times 10^{0}$) &{\bf 3.88}$\pm$0.48 ($\times 10^{-2}$)
    &{\bf 8.23}$\pm$0.35 ($\times 10^{-5}$) &{\bf 3.38}$\pm$0.45 ($\times 10^{-1}$) &{\bf 1.49}$\pm$0.32 ($\times 10^{-2}$)
    &{\bf 5.29}$\pm$0.08 ($\times 10^{-5}$) &{\bf 1.20}$\pm$0.04 ($\times 10^{-1}$) &7.48$\pm$2.53 ($\times 10^{-3}$)
    \\[1pt]
    ENO (fixed) 
    &3.11$\pm$0.44 ($\times 10^{-3}$) &3.62$\pm$0.54 ($\times 10^{1}$) &8.32$\pm$3.14 ($\times 10^{-1}$)
    &1.05$\pm$0.07 ($\times 10^{-4}$) &4.51$\pm$0.26 ($\times 10^{-1}$) &2.42$\pm$0.53 ($\times 10^{-2}$)
    &5.42$\pm$0.14 ($\times 10^{-5}$) &1.42$\pm$0.07 ($\times 10^{-1}$) &{\bf 6.20}$\pm$0.05 ($\times 10^{-3}$)
    \\[1pt]
    Vanilla NO 
    &7.72$\pm$0.77 ($\times 10^{-3}$) &1.32$\pm$0.13 ($\times 10^{2}$) &1.88$\pm$0.47 ($\times 10^{0}$)
    &2.65$\pm$0.39 ($\times 10^{-3}$) &2.59$\pm$0.35 ($\times 10^{1}$) &1.88$\pm$0.31 ($\times 10^{-1}$)
    &6.88$\pm$0.16 ($\times 10^{-5}$) &1.45$\pm$0.03 ($\times 10^{-1}$) &6.33$\pm$0.10 ($\times 10^{-3}$)
    \\[1pt]
    DeepONet 
    &9.18$\pm$0.29 ($\times 10^{-3}$) &1.68$\pm$0.07 ($\times 10^{2}$) &3.60$\pm$0.08 ($\times 10^{0}$)
    &4.75$\pm$0.59 ($\times 10^{-3}$) &4.27$\pm$1.63 ($\times 10^{1}$) &8.04$\pm$2.73 ($\times 10^{-1}$)
    &5.22$\pm$0.59 ($\times 10^{-4}$) &1.45$\pm$0.28 ($\times 10^{0}$) &7.86$\pm$2.71 ($\times 10^{-2}$)
    \\[1pt]
    \bottomrule
\end{tabular}
}
}
\subtable[Cahn–Hilliard equation]{
\scalebox{0.68}{
\begin{tabular}{rrrr|rrr|rrr} \toprule
    &\multicolumn{3}{c|}{$(N_{\rm{x}},N_{\rm{t}})=(10,10)$} 
    &\multicolumn{3}{c|}{$(N_{\rm{x}},N_{\rm{t}})=(15,15)$}
    &\multicolumn{3}{c}{$(N_{\rm{x}},N_{\rm{t}})=(20,20)$}
    \\
    \cmidrule(l){2-4} 
    \cmidrule(l){5-7} 
    \cmidrule(l){8-10}
    &Traj. &Energy &Mass 
    &Traj. &Energy &Mass 
    &Traj. &Energy &Mass\\ \midrule
    ENO 
    &{\bf 7.09}$\pm$0.45 ($\times 10^{-2}$) &{\bf 6.24}$\pm$0.57 ($\times 10^{-5}$) &{\bf 9.98}$\pm$3.93 ($\times 10^{-1}$)
    &{\bf 2.71}$\pm$0.15 ($\times 10^{-3}$) &{\bf 1.64}$\pm$0.28 ($\times 10^{-5}$) &{\bf 2.76}$\pm$0.39 ($\times 10^{-1}$)
    &{\bf 4.51}$\pm$0.46 ($\times 10^{-4}$) &{\bf 7.06}$\pm$1.43 ($\times 10^{-7}$) &{\bf 2.21}$\pm$0.47 ($\times 10^{-2}$)
    \\[1pt]
    ENO (fixed)
    &1.54$\pm$0.54 ($\times 10^{-1}$) &1.46$\pm$1.13 ($\times 10^{-3}$) &1.38$\pm$0.67 ($\times 10^{1}$)
    &1.25$\pm$0.30 ($\times 10^{-2}$) &1.28$\pm$0.48 ($\times 10^{-4}$) &5.44$\pm$2.22 ($\times 10^{-1}$)
    &7.70$\pm$1.31 ($\times 10^{-4}$) &1.59$\pm$0.60 ($\times 10^{-6}$) &2.90$\pm$0.67 ($\times 10^{-2}$)
    \\[1pt]
    Vanilla NO 
    &1.93$\pm$0.17 ($\times 10^{-1}$) &1.92$\pm$0.18 ($\times 10^{-3}$) &4.37$\pm$2.15 ($\times 10^{1}$)
    &3.94$\pm$0.67 ($\times 10^{-2}$) &1.78$\pm$0.40 ($\times 10^{-3}$) &3.00$\pm$1.62 ($\times 10^{0}$)
    &8.71$\pm$0.95 ($\times 10^{-4}$) &1.86$\pm$0.34 ($\times 10^{-6}$) &3.03$\pm$1.49 ($\times 10^{-2}$)
    \\[1pt]
    DeepONet 
    &1.28$\pm$0.18 ($\times 10^{-1}$) &8.95$\pm$2.43 ($\times 10^{-4}$) &3.02$\pm$1.36 ($\times 10^{1}$)
    &1.85$\pm$0.34 ($\times 10^{-2}$) &4.25$\pm$2.79 ($\times 10^{-4}$) &1.18$\pm$0.66 ($\times 10^{0}$)
    &3.39$\pm$0.66 ($\times 10^{-3}$) &1.39$\pm$0.39 ($\times 10^{-5}$) &9.43$\pm$1.80 ($\times 10^{-2}$)
    \\[1pt]
    \bottomrule
\end{tabular}
}
}
\end{sc}
%\end{small}
\end{center}
\end{table}
\end{landscape}

\end{document}